# Halo Reduction in Display Systems through Smoothed Local Histogram Equalization and Human Visual System Modeling

## Prasoon Ambalathankandy, Yafei Ou, Masayuki Ikebe

ikebe@ist.hokudai.ac.jp
Research Center for Integrated Quantum Electronics, Hokkaido University

**ABSTRACT**

*Halo artifacts significantly impact display quality. We propose a method to reduce halos in Local Histogram Equalization (LHE) algorithms by separately addressing dark and light variants. This approach results in visually natural images by exploring the relationship between lateral inhibition and halo artifacts in the human visual system.*

### 1 Introduction

Current display devices, including OLED screens, can be impacted by various artifacts that degrade visual quality, such as ringing, halo, and washed-out color [1]. Halo artifacts significantly detract from the overall visual experience, being prominently visible in images with high contrast transitions. These artifacts not only affect the overall visual quality but also distort sharpness and contrast perception while compromising the accuracy of image details and textures. In response to the challenges associated with improving visual quality, efficient methods have been proposed for edge enhancement and resolution conversion, paying special attention to the mitigation of halo artifacts. The introduced methodology includes a technique for achieving natural-looking edge enhancement using asymmetric weighted halo components [2]. This innovative approach is crucial for producing images that are not only smoother but also visually more natural, all while maintaining low computational costs.

Manufacturers aim to deliver superior display quality by leveraging advanced image processing algorithms and techniques. Attributes such as higher resolutions (4K and 8K), enhanced contrast ratios, color accuracy, refresh rates, and HDR are integral to enhancing the value of display products. In this context, algorithms like Smoothed Local Histogram Equalization (SLHE) are pivotal. SLHE not only improves contrast but also meticulously preserves local details and facilitates adaptive tone mapping, contributing significantly to a visually immersive and realistic viewing experience. Building on the insights and contributions from previous works, this paper proposes a novel approach for halo reduction in display systems. This approach strategically combines SLHE and modeling of the Human Visual System to effectively mitigate halo artifacts, resulting in images that are not only visually pleasing but also more aligned with natural human perception. Our method goes a step further in addressing both dark and light variants of halo artifacts separately through Local Histogram Equalization (LHE), providing a nuanced and comprehensive solution for enhancing display quality.

### 2 Halos in Local Histogram Equalization and Artifact Reduction Method

In smoothed LHE, local windows of the image are analyzed, and histogram equalization is applied to enhance contrast within each window. Moving across the image, the windows encounter transitions between adjacent luminance distributions, creating valleys. When passing through a valley, the input pixel level may deviate from the local mean due to redistribution of pixel intensities during histogram equalization. The target pixel level after equalization can be mapped to a lower or higher level compared to the original input value in the low- or high-level distribution. Consequently, pixel values within the valleys may shift, leading to lower or higher intensities than their original values. These shifts contribute to the formation of halo artifacts.

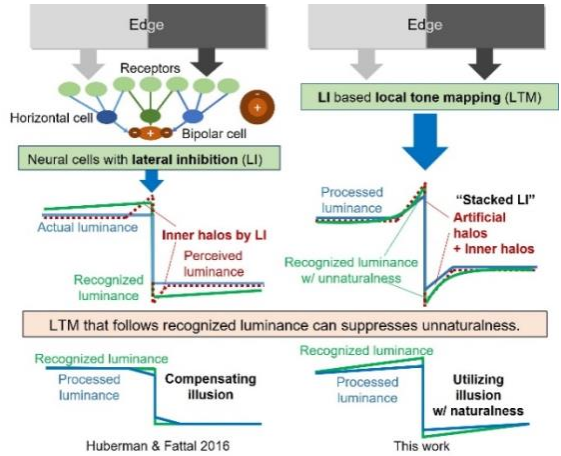

Fig. 1 Comparing HVS and perceived luminance reveals stacked LI effect in tone mapped images, causing unnatural perception.

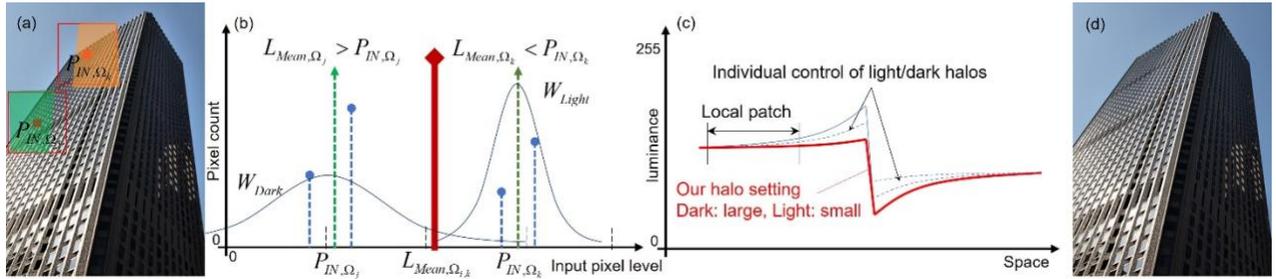

Fig. 2 Principles of the proposed selective dark and light halo control method. Luminance group: Weighting function W, local patch Ω.

Using the image shown in Fig. 1(a) with halo artifacts we will explain our method for controlling the light and dark halos individually in order to generate natural local-edge enhancement using weighted histograms. Our objective is to enhance local edges in the image while minimizing the presence of halos. The input pixel in the polarization luminance distribution is examined to determine whether it belongs to the light or dark luminance group as shown in Fig. 1(b). This determination is made by comparing the local mean (average, represented by $L_{mean}$) with the input pixel level (represented by $P_{IN}$). If the input pixel belongs to the dark luminance group, its weight (represented by $\sigma$) is set to a large value. The weight is determined based on the distance between the local mean and the input pixel level. A larger weight helps in controlling the dark halo. On the other hand, if the input pixel belongs to the light luminance group, its weight is set to a fixed small value.

This small value of $\sigma$ contributes to controlling the light halo. $\sigma_{light}$ represents the σ assigned to the light luminance group, while $\sigma_{dark}$ represents the σ assigned to the dark luminance group. The values of $\sigma_{max}$ and $\sigma_{min}$ play a role in determining the range of σ values. $\sigma_{max}$ represents the maximum value of σ, while $\sigma_{min}$ represents the minimum value of σ. By setting a fixed high value for $\sigma_{min}$ and a linearly changing σ for $\sigma_{dark}$, the treatment of dark halos enables the achievement of natural contrast enhancement. Furthermore, by independently controlling $\sigma_{max}$ and $\sigma_{min}$ as illustrated in Fig. 1(c), it becomes possible to generate images with different artistic or natural characteristics. Figure 1(d) shows the halo suppression achieved using our method. By adjusting these parameters, it is possible to output images with desired visual effects.

$$\sigma_{light} = \sigma_{min} \quad (1)$$

$$\sigma_{dark} = \left(\sigma_{max} - \sigma_{min} \cdot \left[\frac{L_{mean} - P_{IN}}{L_{mean}} + \sigma_{min}\right]\right) \quad (2)$$

## 3 Effects of Light and Dark Halos on Visual Perception

What is the effect of light and dark halos on visual perception? Light/dark halo reduction removes local contrast. The term "halo" is generally used in reference to a light halo. Therefore, we assume that light halos create a feeling of unnaturalness. Kimura and Ikebe evaluated the feelings created by these halos in an experiment with 12 participants [3]. The results of their experiment indicated that using a small value of σ in the range of 50 to 100 is effective in suppressing light halos. On the other hand, it was found that a suitable or appropriate dark halo requires a larger value of σ in the range of 100 to 500. This suggests that dark halos, when present within a certain range, do not create a strong feeling of unnaturalness in visual perception.

The effect of light and dark halos on visual perception can vary as light halos are often associated with a feeling of unnaturalness in an image. These halos can be perceived as a ring or glow around bright regions, causing a loss of detail and affecting the overall image quality [2]. Lateral inhibition is a fundamental process in the HVS that occurs in the retinal cells and neural pathways responsible for visual perception as shown in Fig. (2). It plays a crucial role in enhancing the contrast and edge perception in visual stimuli. By inhibiting the response of neighboring neurons, lateral inhibition helps to sharpen the perception of edges and boundaries between different regions in an image. The relationship between halo artifacts and lateral inhibition lies in the fact that lateral inhibition plays a role in the perception and suppression of halo effects [4].

The visual system's lateral inhibitory mechanisms can contribute to reducing the visibility of halo artifacts by enhancing the contrast at the edges and minimizing the spread of bright or dark regions. Figure 2 illustrates how previous studies have incorporated the principles of lateral inhibition in image processing algorithms [4, 5], and it is possible to reduce the occurrence of halo artifacts. Techniques such as local contrast enhancement, adaptive filtering, or edge-preserving algorithms can leverage the concept of lateral inhibition to suppress halos and improve the overall display quality. LHE filter with a large kernel size acts like an active high pass filter, and it amplifies all the AC frequency components present in the image, particularly those associated with edges and details. This helps in preserving the sharpness and clarity of the edge forms, which contributes to the natural-like feeling of the resulting image.

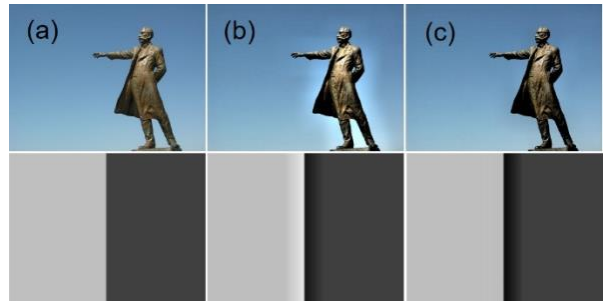

**Fig. 3 Results suppression using the proposed method. (a) Original of halo image, (b) Halo artifacts in processed image, and halo reduced in (c).**

## 4　Conclusions

Image processing algorithms like local contrast enhancement, edge-preserving filtering, and tone mapping utilize lateral inhibition to suppress halos and enhance image quality. Understanding the influence of lateral inhibition from the HVS can enable us to effectively reduce halo artifacts, resulting in improved display quality.